\documentclass{article}

\usepackage{microtype}
\usepackage{graphicx}
\usepackage{subfigure}
\usepackage{booktabs} 
\usepackage{multirow}

\usepackage{hyperref}

 \usepackage[accepted]{icml2025}

\usepackage{amsmath}
\usepackage{amssymb}
\usepackage{mathtools}
\usepackage{amsthm}

\usepackage[capitalize,noabbrev]{cleveref}

\theoremstyle{plain}

\theoremstyle{definition}

\theoremstyle{remark}

\usepackage[textsize=tiny]{todonotes}

\icmltitlerunning{State-Space Models for Tabular Prior-Data Fitted Networks}

\begin{document}

\twocolumn[
\icmltitle{State-Space Models for Tabular Prior-Data Fitted Networks}




\begin{icmlauthorlist}
\icmlauthor{Felix Koch}{thr}
\icmlauthor{Marcel Wever}{luhai}
\icmlauthor{Fabian Raisch}{thr,tum}
\icmlauthor{Benjamin Tischler}{thr}
\end{icmlauthorlist}

\icmlaffiliation{luhai}{L3S Research Center, Leibniz University Hannover, Hannover, Germany}
\icmlaffiliation{thr}{University of Applied Sciences Rosenheim, Rosenheim, Germany}
\icmlaffiliation{tum}{Technical University of Munich, Munich, Germany}

\icmlcorrespondingauthor{Felix Koch}{felix.koch@th-rosenheim.de}
\icmlcorrespondingauthor{Marcel Wever}{m.wever@ai.uni-hannover.de}

\icmlkeywords{Machine Learning, ICML}

\vskip 0.3in
]



\printAffiliationsAndNotice{}  

\begin{abstract}
Recent advancements in foundation models for tabular data, such as TabPFN, demonstrated that pretrained Transformer architectures can approximate Bayesian inference with high predictive performance. However, Transformers suffer from quadratic complexity with respect to sequence length, motivating the exploration of more efficient sequence models.
In this work, we investigate the potential of using Hydra, a bidirectional linear-time structured state space model (SSM), as an alternative to Transformers in TabPFN. 
A key challenge lies in SSM's inherent sensitivity to the order of input tokens -- an undesirable property for tabular datasets where the row order is semantically meaningless. We investigate to what extent a bidirectional approach can preserve efficiency and enable symmetric context aggregation. Our experiments show that this approach reduces the order-dependence, achieving predictive performance competitive to the original TabPFN model.
\end{abstract}

\section{Introduction} 
\label{intro}

Recently, foundation models have shown remarkable performance in tabular classification tasks, particularly in few-shot and small-data regimes \cite{hollmann-iclr23a,zeng2024tabflex,thielmann-arxiv24a}. A prominent example is the Tabular Prior-Data Fitted Network (TabPFN) \cite{hollmann-iclr23a,hollmann-nature25a}, which employs a pretrained Transformer \cite{vaswani-neurips17a} to perform in-context probabilistic inference over labeled tabular datasets. By training on a large corpus of synthetic tasks generated from structural priors, TabPFN can deliver highly accurate predictions in milliseconds without the need for gradient-based adaptation or fine-tuning.

Despite its success, TabPFN inherits the Transformer's computational inefficiencies. The attention mechanism in Transformers has quadratic complexity with respect to the input sequence limiting its scalability. 
Mamba \cite{gu-arxiv23a} presents an alternative that addresses this problem. Mamba is a recently introduced architecture based on structured state space models (SSMs; \citet{gu-iclr22a}). It achieves linear-time sequence processing while retaining the expressivity of the attention mechanism. It has been shown to match or exceed Transformer performance in long-sequence modeling tasks while significantly reducing inference costs \cite{dao-icml24a}. This makes it a promising candidate for scaling TabPFN beyond its limitations on the size of the dataset.
However, substituting the Transformer in TabPFN with Mamba introduces a fundamental issue: Mamba operates as a causal model, meaning its representations are inherently order-sensitive. The authors of \cite{zeng2024tabflex} and \cite{thielmann-arxiv24a} have already noticed this problem. They replaced the Transformer in tabular foundation models with Mamba and observed that Mamba's sensitivity to input order limits its scalability in tabular prediction tasks.
This raises the research question of how to mitigate order sensitivity for SSMs to make them suitable for tabular data. 

To address this issue, we propose to use Hydra \cite{hwang-neurips24a}, a bidirectional extension for SSMs within TabPFN. Hydra uses \textit{quasiseparable matrix mixers} to enable bidirectional sequence modeling. Therewith, we can retain the efficiency of Mamba while allowing symmetric context aggregation across the input and thereby reducing the role of the input order. 
Moreover, we propose repeated context permutations (RCP), invoking Hydra with random input permutations to reduce its order-sensitivity further. In an empirical study, we evaluate substituting the Transformer in TabPFN with Mamba and Hydra.
Our experimental results show that the Hydra-based TabPFN significantly reduces computational and memory complexity, allowing larger inputs while retaining predictive performance similar to the Transformer-based version. We further show that RCPs improve accuracy and align predicted distributions across permutations.
This paper showcases that bidirectional SSMs can also be seen as an alternative to reduce the quadratic complexity alongside approaches with FlashAttention \cite{dao-iclr24a} in \cite{hollmann-nature25a} or Linear Attention \cite{katharopoulos-icml20a} in \cite{zeng2024tabflex}.


\section{Prior-Data Fitted Networks} 
\label{sec:pfn}
Prior-Data Fitted Networks (PFNs, \citet{muller-iclr22a}) are a class of models trained on synthetic tasks sampled from a predefined prior distribution over learning problems called the prior. By learning to predict over such a distribution, these networks approximate Bayesian inference without explicit posterior computation at inference time. Transformers \cite{vaswani-neurips17a} have several benefits to model structured data, making them a commonly used architecture for PFNs. Their core component is the self-attention mechanism, which enables the model to compute interactions between all pairs of inputs in a sequence.

The Tabular Prior-Data Fitted Network (TabPFN) is a Transformer-based instance of PFNs that is especially designed for tabular classification tasks. TabPFN receives an entire dataset as input and classifies it based on the pretrained Transformer model. The model was trained on millions of synthetic classification tasks generated from simple structural causal models and Bayesian neural networks. The offline meta-training enables predictions on a new dataset in a single forward pass, given both the training and test data as input. This produces calibrated outputs within milliseconds, without requiring gradient-based adaptation. However, this performance can only be guaranteed for small datasets due to the limitations of the Transformer architecture. The self-attention operation leads to a quadratic complexity with respect to the input length. The input length corresponds to the number of rows in a tabular setting.


\section{SSMs for Tabular Prior-Fitted Networks}
\label{sec:ssm}

SSMs have emerged as efficient alternatives to Transformers for sequence modeling, especially in tasks involving long-range dependencies. These models are inspired by classical state-space models from control theory and signal processing, where sequences are represented via recurrent updates to (latent) state variables \cite{gu-neurips21a}.

\subsection{Mamba}

One prominent SSM is Mamba \cite{gu-arxiv23a}, which introduces a selective mechanism to update state variables. Thereby, it achieves linear-time inference (cf. Figure \ref{fig:tabpfninput}) while matching or exceeding Transformer performance in several domains \cite{gu-arxiv23a}. For this reason, we are testing Mamba as an alternative for the Transformer within TabPFN. For the implementation of Mamba, we interpret the rows of the table as a sequence of classification examples.

Like other SSMs, Mamba can be expressed as \textit{semiseparable matrix multipliers} \cite{dao-icml24a}, enabling a hardware-efficient implementation. However, most SSMs, including Mamba, are causal by design: they process sequences in a forward (autoregressive) manner.

This may present a problem in tabular data, where the order of rows has no meaning and thus there exists no natural order. Also, Mamba scans in an unidirectional way, making it possible to draw causal connections from only one direction. This may lower in-context learning capabilities for tabular tasks, which was also concluded by \cite{zeng2024tabflex}.

\subsection{Hydra}

\begin{figure}[t]
    \centering
    \includegraphics[width=\columnwidth,trim={0.7cm 0.4cm 0.8cm 0},clip]{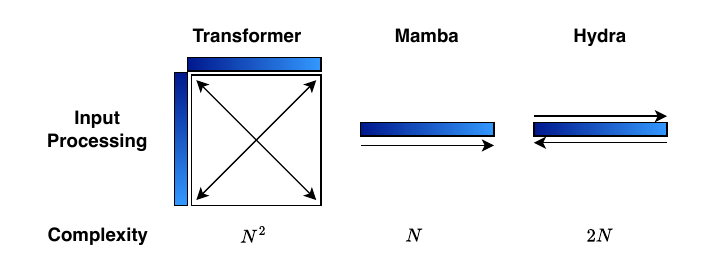}
    \caption{Comparison of different backbone architectures for TabPFN based on their input processing. Transformers have quadratic complexity, whereas Mamba offers linear-time processing, and Hydra adds bidirectional capabilities.}
    \label{fig:tabpfninput}
\end{figure}

To address the issues of Mamba, we additionally consider Hydra \cite{hwang-neurips24a} as another alternative for the Transformer. Hydra is a bidirectional extension of Mamba that enables efficient processing of sequences using \textit{quasiseparable matrix mixers}. Hydra retains the linear-time benefits of Mamba while allowing the model to be more agnostic of the order, which is essential for principled inference on tabular datasets. An overview of the different backbone architectures is provided in \cref{fig:tabpfninput}.

To replace the Transformer in TabPFN with Hydra, we make the following modifications:

\textbf{Backbone Replacement} The Transformer encoder is replaced with a stack of Hydra layers. Each layer consists of a bidirectional state-space mixing. This is followed by feedforward transformations, closely mirroring the Transformer block structure but with linear-time complexity.

\textbf{Embedding Format} We retain the original data embedding strategy, where each input is represented as a concatenation of feature values and a class label. As in TabPFN, inference involves marginalizing over all possible label assignments for unlabeled data.

\textbf{Parameter Compatibility} Due to architectural differences, Hydra-based TabPFN requires retraining on the synthetic task distribution used for the original TabPFN. However, the training pipeline remains unchanged aside from the swap of the backbone.

\subsection{Repeated Context Permutations}
\label{sec:rcp}

SSMs are inherently dependent on the input sequence order, distinguishing them from Transformers, which are intrinsically positionally invariant. This limits the application of SSMs to tabular data, mainly because all training examples in one inference are not considered equally important. To decrease the sensitivity of SSM-based PFNs on the sequence order of the rows of a given dataset, we integrate row-wise \textit{repeated context permutations} (RCP) into the inference. This approach reduces the dependency of the results on row ordering by predicting $r$ times with a shuffled context and averaging the predicted probability distributions. RCPs linearly increase the inference time by a factor of $r$ (see Algorithm \ref{alg:pooling}). To assess the benefit of RCPs for SSM-based TabPFN, we include an ablation study in Section \ref{sec:results}.

\begin{algorithm}[tb]
   \caption{RCP for Tabular PFN}
   \label{alg:pooling}
   \begin{algorithmic}
       \STATE {\bfseries Input:} Number of permutations $r$, context $D$,  $x_{\text{test}}$
       \STATE {\bfseries Output:} Predicted class values
       \STATE Initialize an empty list: $outputs \gets [\:]$
       \FOR{$i = 1$ {\bfseries to} $r$}
            \STATE Shuffle rows of $D$: $D_p \gets \text{shuffle}(D)$
            \STATE Concatenate $x_{\text{test}}$ to $D_p$: $D_{\text{in}} \gets D_p \cup x_{\text{test}}$
            \STATE Predict: $outputs[i] \gets \text{PFN.predict}(D_{\text{in}})$
       \ENDFOR
       \STATE \textbf{return} average of $outputs$
   \end{algorithmic}
\end{algorithm} 


\section{Evaluation}
\label{sec:results}

In this section, we compare our implementation of Mamba and Hydra with the state-of-the-art Transformer model. 

\subsection{Experiment Setup}
\label{sec:setup}

For evaluation, we employ publicly available datasets from OpenML \cite{vanschoren-sigkdd14a}. All models are evaluated on the multiclass classification datasets from the OpenML CC-18 benchmarking suite \cite{bischl-arxiv17a}. More specifically, these datasets were filtered to contain $\leq 2000$ rows, $\leq 100$ features, and $\leq 10$ classes beforehand to meet the constraints of TabPFN. This results in a total of 30 analyzed datasets. Each dataset is randomly split into training and test sets 16 times to account for variance in observations. 
We evaluated each model in the test set and reported the mean and standard error across all splits.
For more details on our experiments, we refer to Appendix~\ref{apx:experimentalsetup}. The code is publicly available on \href{https://github.com/felixmkoch/Structured-State-Space-Models-for-PFNs}{GitHub}.


As stated in Section \ref{sec:rcp}, SSMs are inherently dependent on the order of the input sequence. To measure the influence of this order on the predictions, we employ the KL divergence: 
\begin{equation*}
    D_{KL}(P || Q) = \sum_i P_i log(\frac{P_i}{Q_i}) \,\, .
    \label{eq:kldiv}
\end{equation*}
We compare two predictions with shuffled contexts and quantify the impact of the row ordering on the resulting output. The KL divergence then provides an indication of the dependency. A higher value for the KL divergence indicates a higher entropy and thus a greater dependence on the order of the rows in the table for the model.

\subsection{Results}

In the following, we will present three experiments that compare Mamba and Hydra as replacements for the Transformer within TabPFN. First, we will asses the inference times, as these are the main motivation for using SSMs versus the Transformer model. Second, we will evaluate the performance of the models on the datasets. This provides a raw understanding of the performance loss when using SSMs compared to the Transformer. Lastly, we will evaluate the dependency on the order of the input sequence from reasons stated in Sections \ref{sec:rcp} and \ref{sec:setup}.

\begin{figure}[t]
    \vskip 0.2in
        \begin{center}
            \centerline{\includegraphics[width=.95\columnwidth]{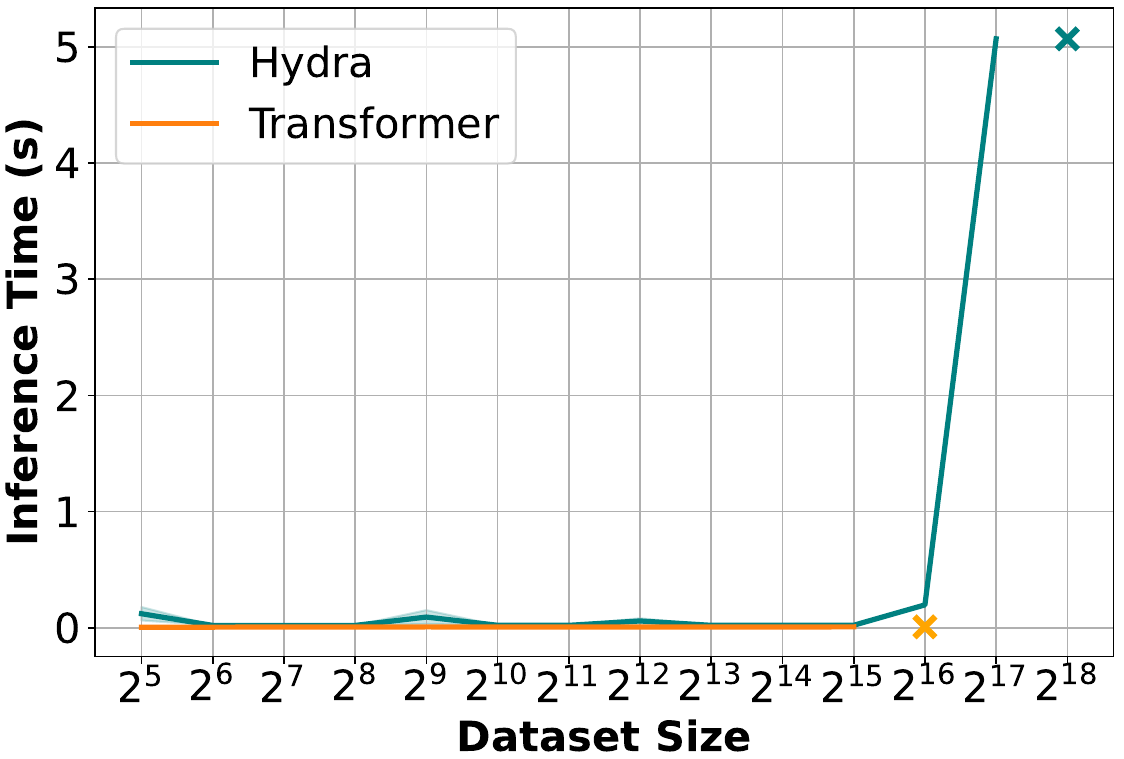}}
            \caption{Mean inference time (standard error as shaded areas) for increasing input dataset sizes, ranging from $2^5$ to $2^{17}$, comparing Hydra- and Transformer-based TabPFN, on an H100 with 80GB VRAM. The Transformer allows for input sizes of up to $2^{15}$, and Hydra up to $2^{17}$}
            \label{fig:inference_time}
        \end{center}
    \vskip -0.2in
\end{figure}

First, we compare inference times of Transformer-based TabPFN and its Hydra-backed variant across varying input sizes. Mamba exhibited similar behavior to Hydra; therefore, its results are omitted for clarity. As shown in \cref{fig:inference_time}, the mean inference time grows with dataset size. For the Transformer, inference fails at $2^{16}$ rows, as the space requirements for the quadratic self-attention matrix exceed the available 80GB VRAM. Hydra only fails at $2^{18}$ rows, i.e., a two orders of magnitude larger dataset. However, it \textit{does not} exceed the available memory yet, but PyTorch’s 32-bit indexing limit, which is a software limitation but not a hardware one. Note that the scalability of Transformer can be improved through, e.g., FlashAttention, reducing space requirements by a constant factor of 20 \cite{dao-neurips22a}.

\begin{figure*}[t]
    \centering
    \includegraphics[trim={0.0cm 0.5cm 0.0cm 0},clip,width=.75\textwidth]{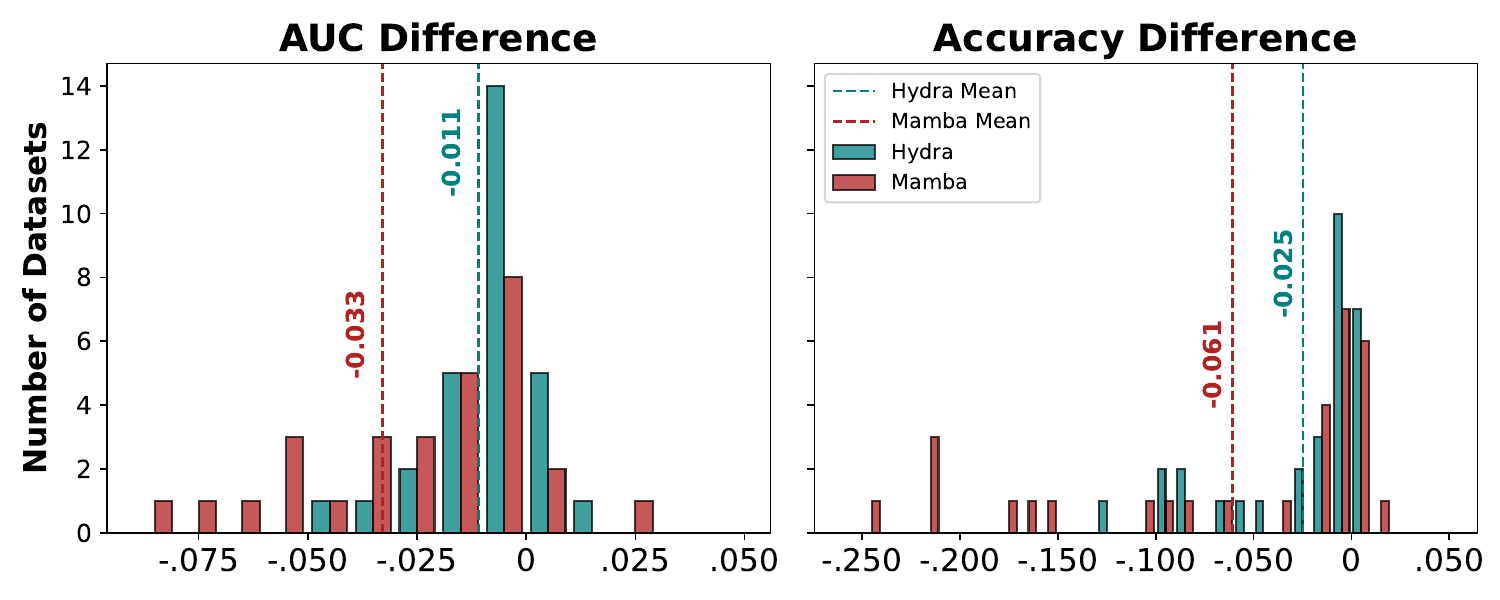}
    \caption{Comparison of the performance of Mamba and Hydra models with the Transformer as a baseline}
    \label{fig:performanceresults}
\end{figure*}

\begin{figure}[t]
    \vskip 0.2in
        \begin{center}
            \centerline{\includegraphics[width=\columnwidth]{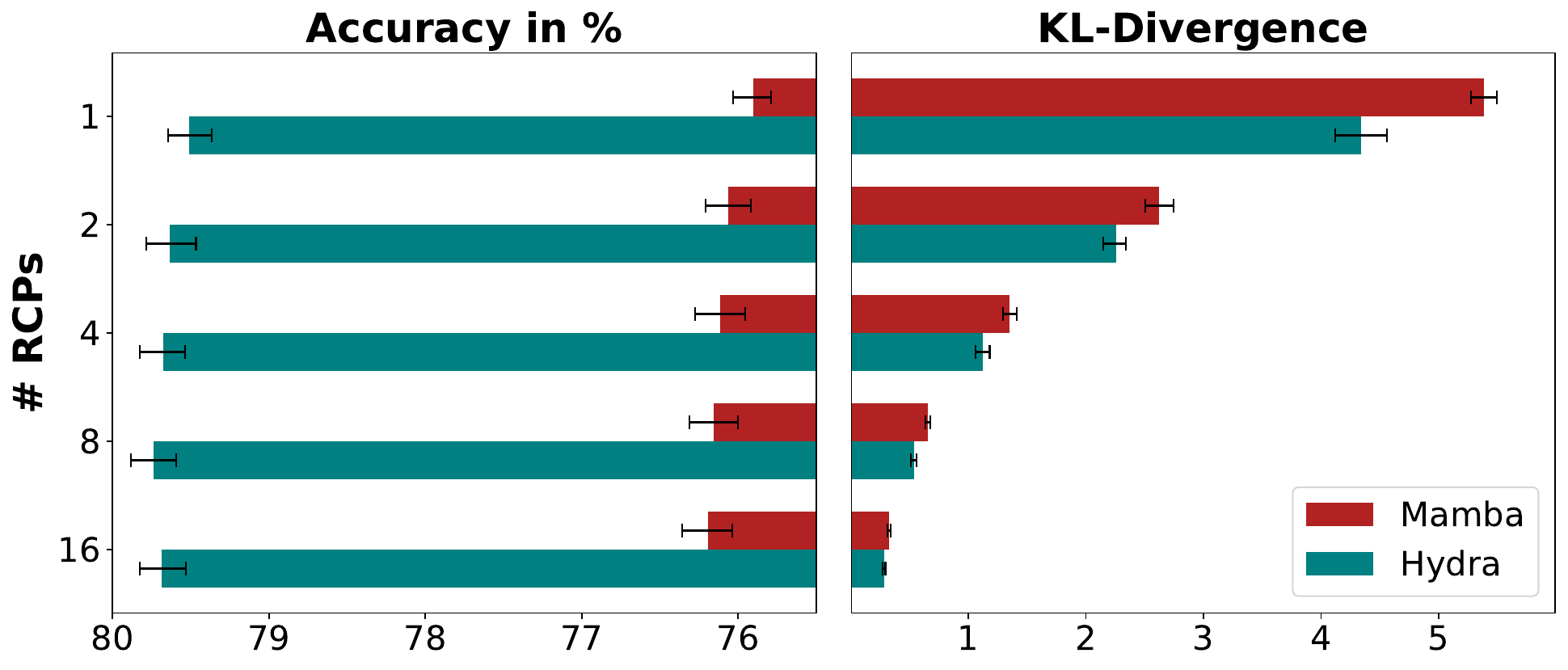}}
            \caption{Effect of Repeated Context Permutations on the KL divergence and the accuracy in TabPFN with Hydra and Mamba as a backbone model.}
            \label{fig:kldivbars}
        \end{center}
    \vskip -0.2in
\end{figure}
Second, we compare the performance of Mamba and Hydra in terms of accuracy and AUC OvO in Figure \ref{fig:performanceresults} (detailed results in Appendix \ref{app:additional_results1}). Here, we calculate the performance difference for each dataset between the corresponding SSM and the Transformer. 
The results show that Mamba exhibits higher variance than Hydra in both AUC and accuracy. On average, Hydra achieves 3.6\% higher accuracy than Mamba, showing a clear advantage of the bidirectional approach over the unidirectional approach. Furthermore, Hydra achieves an average difference of 1.1\%, indicating that its performance is quite close to that of the Transformer. Notably, Hydra also attains the best performance on some of the datasets.

Third, we investigate the influence of the number of RCPs on our results. Figure \ref{fig:kldivbars} presents the effects of the RCPs on the KL divergence and the quality of the prediction in terms of accuracy. In addition to the average accuracy, 95\% confidence intervals were added to show the variance over the splits (cf. \ref{sec:setup}). As can be seen, order sensitivity is decreased as the number of RCPs increases. Also, with an increasing number of RCPs, the accuracy increases to a certain extent, which can be attributed to outliers based on a disadvantageous ordering of table rows being averaged out. This effect on the accuracy is comparably small compared to the relatively large variance. We skip AUC here because no significant improvements can be observed for this metric.


\section{Conclusion} 
This paper introduced the use of State-Space Models for tabular foundation models. Mamba and Hydra are analyzed as Tabular Prior-Data Fitted Networks and compared to the state of the art, i.e., the Transformer architecture. The results show that our approach can deal with larger input sizes due to its linear complexity as opposed to the quadratic complexity of Transformers while maintaining competitive performance. In contrast to other approaches that aimed to enable large table analysis for tabular PFNs, our method directly addresses the quadratic complexity of the underlying Transformer architecture, thereby targeting the root cause of scalability limitations rather than just fixing symptoms.

Comparative analyses revealed that the bidirectional approach Hydra performed better on average than Mamba, which parsed inputs unidirectionally. Additionally, we found the use of multiple Repeated Context Permutations to be useful. These findings suggest that Hydra-based PFNs, combined with Repeated Context Permutations, hold promise for analyzing large tabular datasets, and we hope they inspire further work in this direction.

For the number of rows in the context, we limited ourselves to 1000, conforming to \citet{hollmann-iclr23a}. The most promising use case and future direction is to test SSMs as tabular PFNs for longer contexts (for example, $>10$k rows, the current limit for TabPFNv2 in \cite{hollmann-nature25a}).
In addition, future work may further investigate how to mitigate the impact of the row order on the prediction of an SSM-based tabular PFN.
We support the assumption proposed by \cite{thielmann-arxiv24a} that certain row orderings of the context may enhance the performance of SSMs.
Moreover, such an optimal ordering may differ for unidirectional and bidirectional SSMs.

\bibliography{strings, lib, proc, refs}
\bibliographystyle{icml2025}

\newpage
\appendix
\onecolumn
\section{Experimental Setups}
\label{apx:experimentalsetup}

This section further describes how the above-mentioned experiments were conducted and which hardware was used.

\subsection{Training}
Training was done on one Nvidia A40 GPU with 48GB of memory. We trained the models based on the Transformer for 48 hours, Mamba for 52 hours, and Hydra for 134 hours. Each training run independently sampled datasets from the same Prior. Thus, it was assumed that the training environment for all models is similar due to the same underlying distribution.

\subsection{Validation}
The best model was selected based on the best score on the validation set. For this, we used the same datasets as \cite{hollmann-iclr23a} listed in Table \ref{tab:validationdataset}. We excluded two datasets due to their overlap with the OpenML CC-18 benchmarks used for testing. \\

\begin{table}[h]
\label{tab:validationdataset}
    \caption{Validation Set from OpenML}
    \begin{center}
        \begin{footnotesize}
            \begin{tabular}{l l l l l } 
                \hline
                \textbf{Dataset-ID} & \textbf{Dataset Name} & \textbf{\# Instances} & \textbf{\# Features} & \textbf{\# Classes} \\
                \hline
                13 & breast-cancer & 286 & 9 & 2 \\
                43 & haberman & 306 & 3 & 2 \\
                59 & ionosphere & 351 & 34 & 2 \\
                1498 & sa-heart & 400 & 9 & 2 \\
                40710 & cleave & 303 & 13 & 2 \\

                \hline
            \end{tabular}
            \vspace{15pt}
        \end{footnotesize}
    \end{center}
\end{table}

\subsection{Testing Input Sequence Lengths}
We used a node equipped with an NVIDIA H100 80GB, 64 CPU Cores, and 64GB RAM to test increasing input sequence lengths. For testing, we repeated the executions of Hydra- and Transformer-based TabPFN 10 times. Furthermore, instead of specific datasets, we generated random tensors with 99 feature columns and the number of rows ranging from $2^5$ to $2^{18}$.

\subsection{Hyperparameter}
\label{appx:hyperparam}
A thorough hyperparameter optimization was not feasible due to long training times. We found that the number of steps per epoch and the batch size proposed in \cite{hollmann-iclr23a} lead to catastrophic forgetting effects in both Mamba and Hydra. Therefore, we changed those values to avoid such issues. By default, we used double the number of Transformer encoder layers as SSM blocks as suggested by \cite{gu-arxiv23a}. Table \ref{tab:hyperparams} lists all hyperparameters used for training.

\begin{table}[h]
\label{tab:hyperparams}
    \caption{Hyperparameter Table for Training}
    \begin{center}
        \begin{footnotesize}
            \begin{tabular}{l l l l} 
                \hline
                \textbf{Hyperparameter} & \textbf{Mamba} & \textbf{Transformer} & \textbf{Hydra} \\ [0.5ex] 
                \hline
                Learning Rate & 0.0001 & 0.0001 & 0.0001 \\
                Dropout & 0.0 & 0.0 & 0.0 \\
                Batch Size & 512 & 64 & 128\\
                Steps per Epoch & 16 & 1024 & 64 \\
                Aggregate k Gradients & 8 & 8 & 8 \\
                \hline
                Embedding Size & 1024 & 512 & 1024 \\
                Hidden Size & 1024 & 1024 & 1024 \\
                Number of Layers & 24 & 12 & 24 \\
                Number of Heads & - & 4 & -\\
                Recompute Attention & - & True & - \\
                Use AMP & True & True & True \\
                Optimizer & AdamW & AdamW & AdamW  \\
                \hline
            \end{tabular}
            \vspace{15pt} 
        \end{footnotesize}
    \end{center}
\end{table}

\subsection{OpenML CC-18 Datasets}  

We used the OpenML CC-18 benchmark datasets, filtered for TabPFN limitations \cite{hollmann-iclr23a}. Table \ref{tab:openmldatasets} lists each dataset with its respective characteristics.

\begin{table}[h]
\label{tab:openmldatasets}
    \caption{OpenML CC-18 Datasets Table, filtered on TabPFN limitations}
    \begin{center}
        \begin{footnotesize}
            \begin{tabular}{l l | l l l l l} 
                \hline
                \textbf{Dataset-ID} & \textbf{Dataset Name} & \textbf{\# Instances} & \textbf{\# Features} & \textbf{\# Classes} & \textbf{\# NaNs} & \textbf{Size Min. Class}\\
                \hline
                11 & balance-scale & 625 & 5 & 3 & 0 & 49 \\
                14 & mfeat-fourier & 2000 & 77 & 10 & 0 & 200 \\
                15 & breast-w & 699 & 10 & 2 & 16 & 241 \\
                16 & mfeat-karhunen & 2000 & 65 & 10 & 0 & 200 \\
                18 & mfeat-morphological & 2000 & 7 & 10 & 0 & 200 \\
                22 & mfeat-zernike & 2000 & 48 & 10 & 0 & 200 \\
                23 & cmc & 1473 & 10 & 3 & 0 & 333 \\
                29 & credit-approval & 690 & 16 & 2 & 67 & 307 \\
                31 & credit-g & 1000 & 21 & 2 & 0 & 300 \\
                37 & diabetes & 768 & 9 & 2 & 0 & 268 \\
                50 & tic-tac-toe & 958 & 10 & 2 & 0 & 332 \\
                54 & vehicle & 846 & 19 & 4 & 0 & 199 \\
                188 & eucalyptus & 736 & 20 & 5 & 448 & 105 \\
                458 & analcatdata\_authorship & 841 & 71 & 4 & 0 & 55 \\
                469 & analcatdata\_dmft & 797 & 5 & 6 & 0 & 123 \\
                1049 & pc4 & 1458 & 38 & 2 & 0 & 178 \\
                1050 & pc3 & 1563 & 38 & 2 & 0 & 160 \\
                1063 & kc2 & 522 & 22 & 2 & 0 & 107 \\
                1068 & pc1 & 1109 & 22 & 2 & 0 & 77 \\
                1462 & banknote-authentication & 1372 & 5 & 2 & 0 & 610 \\
                1464 & blood-transfusion-service-center & 748 & 5 & 2 & 0 & 178 \\
                1480 & ilpd & 583 & 11 & 2 & 0 & 167 \\
                1494 & qsar-biodeg & 1055 & 42 & 2 & 0 & 356 \\
                1510 & wdbc & 569 & 31 & 2 & 0 & 212 \\
                6332 & cylinder-bands & 540 & 40 & 2 & 999 & 228 \\
                23381 & dresses-sales & 500 & 13 & 2 & 835 & 210 \\
                40966 & MiceProtein & 1080 & 82 & 8 & 1396 & 105 \\
                40975 & car & 1728 & 7 & 4 & 0 & 65 \\
                40982 & steel-plates-fault & 1941 & 28 & 7 & 0 & 55 \\
                40994 & climate-model-simulation-crashes & 540 & 21 & 2 & 0 & 46 \\
               
                \hline
            \end{tabular}
            \vspace{15pt}
        \end{footnotesize}
    \end{center}
\end{table}

\section{Additional Results}

Here, we provide additional results that put the performance of our approaches more into context. The environment for those results was the same as in the other ones in this paper.

\subsection{Detailed Performance of SSMs and TabPFN over OpenML Datasets}
\label{app:additional_results1}

As in Section \ref{sec:results}, we compare both SSM architectures with the Transformer-based TabPFN on the OpenML CC-18 benchmark. Model evaluation was performed on each resulting test set, where the average and the standard deviation of all 16 splits were reported. In Table \ref{tab:results:metrics}, bold values denote the best performance regarding dataset ID and the specific metric. The second best value is underlined if the result is not significantly improved (meaning $p \nless 0.05$), determined by a two-sided Wilcoxon signed rank test.

\begin{table*}[t]
    \centering
    \caption{Results on the OpenML CC-18 dataset with constraints}
    \label{tab:results:metrics}
    \begin{footnotesize}
        \begin{tabular}{l | l@{\hspace{18pt}}l@{\hspace{18pt}}l | l@{\hspace{18pt}}l@{\hspace{18pt}}l}
            \toprule
            \multirow{2}{*}{DID} & \multicolumn{3}{c|}{AUC OVO} & \multicolumn{3}{c}{Accuracy} \\
            \cmidrule(lr){2-4} \cmidrule(lr){5-7}
             & Transformer & Mamba & Hydra & Transformer & Mamba & Hydra \\
            \midrule
            11 & \textbf{0.9977} $\pm$ .003 & 0.9648 $\pm$ .008 & 0.9951 $\pm$ .004 & \textbf{0.9824} $\pm$ .010 & 0.8968 $\pm$ .013 & 0.9702 $\pm$ .015\\
            14 & \textbf{0.9753} $\pm$ .003 & 0.9274 $\pm$ .007 & 0.9557 $\pm$ .005 & \textbf{0.7925} $\pm$ .017 & 0.5751 $\pm$ .020 & 0.7002 $\pm$ .018\\
            15 & 0.9942 $\pm$ .002 & 0.9934 $\pm$ .002 & \textbf{0.9946} $\pm$ .001 & 0.9698 $\pm$ .004 & \textbf{0.9732} $\pm$ .003 & 0.9679 $\pm$ .004\\
            16 & \textbf{0.9969} $\pm$ .001 & 0.9657 $\pm$ .006 & 0.9857 $\pm$ .003 & \textbf{0.9418} $\pm$ .009 & 0.7708 $\pm$ .028 & 0.8604 $\pm$ .019\\
            18 & \textbf{0.9635} $\pm$ .003 & 0.9461 $\pm$ .003 & 0.9574 $\pm$ .003 & \textbf{0.7361} $\pm$ .014 & 0.6422 $\pm$ .018 & 0.6949 $\pm$ .017\\
            22 & \textbf{0.9819} $\pm$ .001 & 0.9270 $\pm$ .010 & 0.9697 $\pm$ .003 & \textbf{0.8218} $\pm$ .015 & 0.6029 $\pm$ .031 & 0.7352 $\pm$ .021\\
            23 & \underline{0.7187} $\pm$ .015 & 0.7027 $\pm$ .016 & \textbf{0.7213} $\pm$ .016 & \underline{0.5414} $\pm$ .016 & 0.5243 $\pm$ .019 & \textbf{0.5422} $\pm$ .015\\
            29 & \textbf{0.9308} $\pm$ .016 & 0.9287 $\pm$ .016 & 0.9288 $\pm$ .017 & \textbf{0.8677} $\pm$ .019 & 0.8664 $\pm$ .019 & \underline{0.8673} $\pm$ .018\\
            31 & \textbf{0.7971} $\pm$ .014 & 0.7846 $\pm$ .015 & 0.7911 $\pm$ .015 & \textbf{0.7631} $\pm$ .013 & 0.7511 $\pm$ .009 & \underline{0.7625} $\pm$ .012\\
            37 & \underline{0.8356} $\pm$ .011 & 0.8342 $\pm$ .010 & \textbf{0.8369} $\pm$ .011 & 0.7676 $\pm$ .011 & \underline{0.7703} $\pm$ .011 & \textbf{0.7712} $\pm$ .008\\
            50 & \textbf{0.9728} $\pm$ .010 & 0.6739 $\pm$ .022 & 0.8764 $\pm$ .022 & \textbf{0.9087} $\pm$ .013 & 0.6901 $\pm$ .019 & 0.8184 $\pm$ .019\\
            54 & \textbf{0.9580} $\pm$ .003 & 0.8939 $\pm$ .007 & 0.9293 $\pm$ .006 & \textbf{0.8156} $\pm$ .014 & 0.6622 $\pm$ .018 & 0.7459 $\pm$ .015\\
            188 & \textbf{0.9143} $\pm$ .006 & 0.8776 $\pm$ .008 & 0.9049 $\pm$ .005 & \textbf{0.6539} $\pm$ .020 & 0.5906 $\pm$ .021 & 0.6361 $\pm$ .015\\
            458 & \textbf{1.0000} $\pm$ .0 & 0.9997 $\pm$ .0 & 0.9999 $\pm$ .0 & \textbf{0.9979} $\pm$ .002 & 0.9905 $\pm$ .004 & 0.9954 $\pm$ .003\\
            469 & \textbf{0.5651} $\pm$ .015 & 0.5500 $\pm$ .012 & \underline{0.5649} $\pm$ .017 & 0.1891 $\pm$ .014 & \textbf{0.1916} $\pm$ .015 & \underline{0.1908} $\pm$ .015\\
            1049 & \textbf{0.9326} $\pm$ .011 & 0.9044 $\pm$ .014 & 0.9248 $\pm$ .010 & \textbf{0.9021} $\pm$ .013 & 0.8911 $\pm$ .015 & \underline{0.9008} $\pm$ .011\\
            1050 & \textbf{0.8298} $\pm$ .023 & 0.8080 $\pm$ .022 & 0.8194 $\pm$ .021 & 0.8982 $\pm$ .010 & \underline{0.8987} $\pm$ .009 & \textbf{0.8992} $\pm$ .009\\
            1063 & \textbf{0.8442} $\pm$ .025 & \underline{0.8420} $\pm$ .026 & 0.8412 $\pm$ .025 & \underline{0.8379} $\pm$ .018 & 0.8372 $\pm$ .015 & \textbf{0.8408} $\pm$ .015\\
            1068 & \textbf{0.8851} $\pm$ .027 & 0.8266 $\pm$ .021 & 0.8447 $\pm$ .027 & 0.9299 $\pm$ .011 & \underline{0.9301} $\pm$ .011 & \textbf{0.9301} $\pm$ .011\\
            1462 & \textbf{1.0000} $\pm$ .0 & \textbf{1.0000} $\pm$ .0 & \textbf{1.0000} $\pm$ .0 & \textbf{0.9999} $\pm$ .0 & 0.9982 $\pm$ .002 & 0.9993 $\pm$ .001\\
            1464 & 0.7552 $\pm$ .021 & \underline{0.7550} $\pm$ .024 & \textbf{0.7562} $\pm$ .024 & \textbf{0.7883} $\pm$ .013 & \underline{0.7871} $\pm$ .015 & 0.7843 $\pm$ .010\\
            1480 & \underline{0.7412} $\pm$ .027 & \textbf{0.7429} $\pm$ .022 & 0.7374 $\pm$ .023 & \underline{0.7075} $\pm$ .028 & \textbf{0.7135} $\pm$ .019 & 0.7047 $\pm$ .024\\
            1494 & \textbf{0.9324} $\pm$ .007 & 0.9165 $\pm$ .007 & 0.9229 $\pm$ .007 & \textbf{0.8784} $\pm$ .012 & 0.8600 $\pm$ .011 & 0.8656 $\pm$ .014\\
            1510 & \textbf{0.9963} $\pm$ .002 & \underline{0.9962} $\pm$ .003 & 0.9960 $\pm$ .003 & \textbf{0.9769} $\pm$ .009 & 0.9727 $\pm$ .007 & 0.9736 $\pm$ .009\\
            6332 & \textbf{0.7755} $\pm$ .035 & 0.7200 $\pm$ .034 & 0.7430 $\pm$ .034 & \textbf{0.7314} $\pm$ .037 & 0.6916 $\pm$ .024 & 0.7097 $\pm$ .043\\
            23381 & 0.4637 $\pm$ .069 & \textbf{0.4885} $\pm$ .071 & \underline{0.4813} $\pm$ .061 & 0.5587 $\pm$ .041 & \textbf{0.5778} $\pm$ .030 & 0.5625 $\pm$ .030\\
            40966 & \textbf{1.0000} $\pm$ .0 & 0.9716 $\pm$ .008 & 0.9908 $\pm$ .003 & \textbf{0.9950} $\pm$ .007 & 0.7534 $\pm$ .046 & 0.8700 $\pm$ .025\\
            40975 & \textbf{0.9880} $\pm$ .006 & 0.9057 $\pm$ .015 & 0.9775 $\pm$ .010 & \textbf{0.9597} $\pm$ .009 & 0.8523 $\pm$ .012 & 0.9336 $\pm$ .015\\
            40982 & \textbf{0.9547} $\pm$ .007 & 0.8758 $\pm$ .010 & 0.9280 $\pm$ .008 & \textbf{0.7354} $\pm$ .017 & 0.5685 $\pm$ .025 & 0.6757 $\pm$ .023\\
            40994 & \textbf{0.9444} $\pm$ .014 & 0.9352 $\pm$ .015 & 0.9428 $\pm$ .014 & \textbf{0.9475} $\pm$ .014 & 0.9412 $\pm$ .015 & \underline{0.9472} $\pm$ .014\\

            \bottomrule
        \end{tabular}
    \end{footnotesize}
\end{table*}

\subsection{KL-Divergence over RPC}

\begin{figure}[htbp]
    \centering
    \subfigure[Hydra]{
        \includegraphics[width=0.45\textwidth]{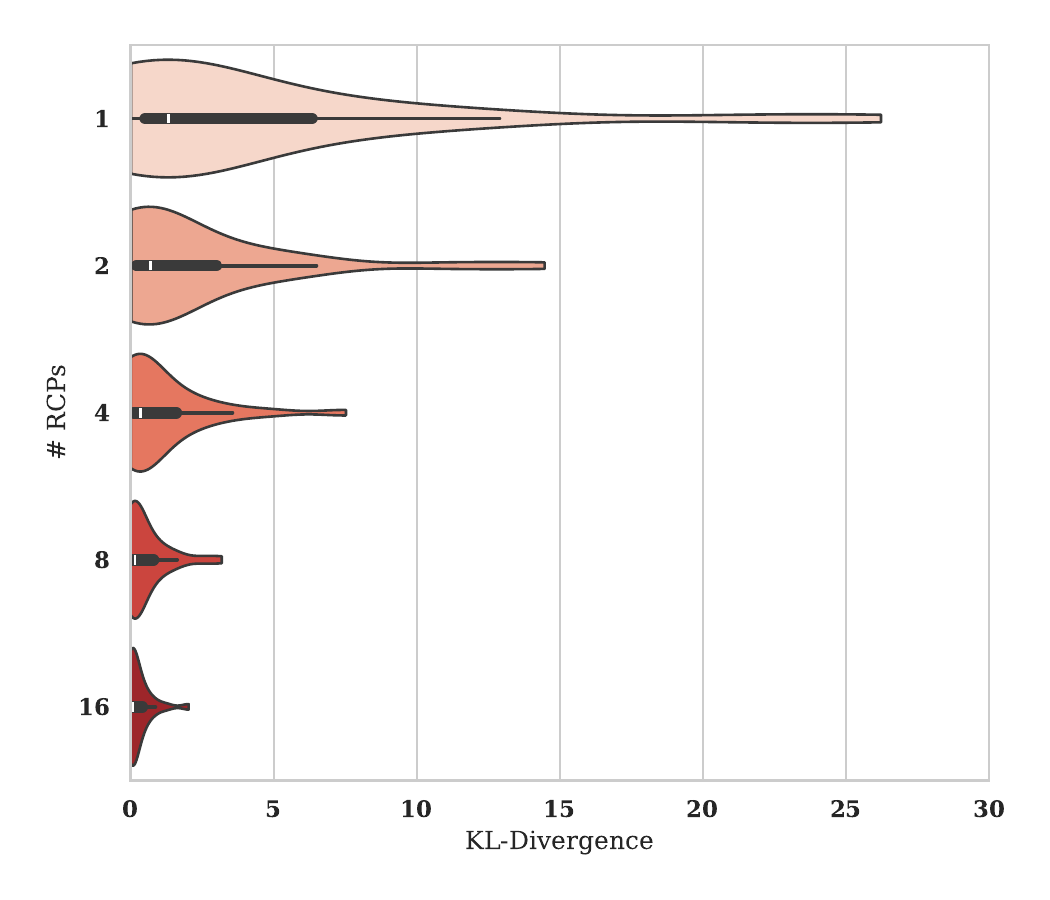}
        \label{fig:violine_hydra}
    }
    \hfill
    \subfigure[Mamba]{
        \includegraphics[width=0.45\textwidth]{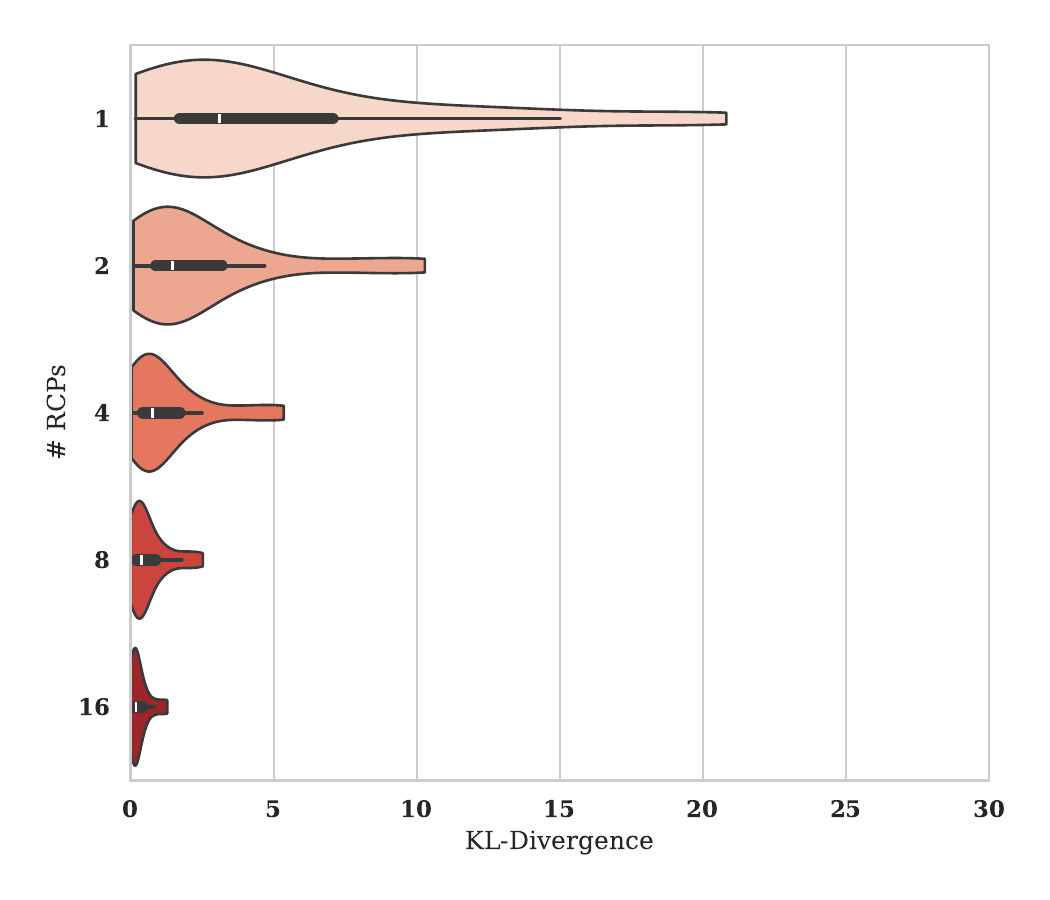}
        \label{fig:violine_mamba}
    }
    \caption{Impact of RPC on KL-divergence of SSM-based PFN class values}
    \label{fig:kl_ssm}
\end{figure}

In Section \ref{sec:results}, we thematized how RCP impacted the KL divergence of the Mamba- and Hydra-based PFNs by averaging the values for the 30 datasets. Figure \ref{fig:kl_ssm} highlights the distribution of the KL-divergence for the datasets.

\end{document}